%% file: main.tex
\documentclass[sigconf]{acmart}

\AtBeginDocument{%
  \providecommand\BibTeX{{%
    \normalfont B\kern-0.5em{\scshape i\kern-0.25em b}\kern-0.8em\TeX}}}

\copyrightyear{2024}
\acmYear{2024}
\setcopyright{acmlicensed}\acmConference[WWW '24 Companion]{Companion Proceedings of the ACM Web Conference 2024}{May 13--17, 2024}{Singapore, Singapore}
\acmBooktitle{Companion Proceedings of the ACM Web Conference 2024 (WWW '24 Companion), May 13--17, 2024, Singapore, Singapore}
\acmDOI{10.1145/3589335.3648321}
\acmISBN{979-8-4007-0172-6/24/05}

\acmPrice{}
\usepackage{graphicx} 
\usepackage{enumitem}
\usepackage{booktabs} 
\usepackage{siunitx} 
\usepackage{makecell} 

\acmSubmissionID{13303.65}

\usepackage{algpseudocode}
\usepackage[linesnumbered,ruled,vlined]{algorithm2e}
\begin{document}

\title{Large Multimodal Model Compression via Iterative Efficient Pruning and Distillation}


\author{Maolin Wang}
\authornote{Both authors contributed equally to this research.}
\authornote{All work performed during the internship at Ant Group}
\affiliation{%
  \institution{City University of Hong Kong}
   \city{Hong Kong SAR}
   \country{China}
  }
\email{morin.wang@my.cityu.edu.hk}

\author{Yao Zhao}
\authornotemark[1]
\affiliation{%
  \institution{AntGroup}
  \city{Hangzhou}
  \country{China}
  }
\email{nanxiao.zy@antgroup.com}

\author{Jiajia Liu}
\affiliation{%
  \institution{AntGroup}
  \city{Hangzhou}
  \country{China}
  }
\email{lekun.ljj@antgroup.com}

\author{Jingdong Chen}
\affiliation{%
  \institution{AntGroup}
  \city{Hangzhou}
  \country{China}
  }
\email{jingdongchen.cjd@antgroup.com}

\author{Chenyi Zhuang}
\authornotemark[4]
\affiliation{%
  \institution{AntGroup}
  \city{Hangzhou}
  \country{China}
  }
\email{chenyi.zcy@antgroup.com}

\author{Jinjie Gu}
\affiliation{%
  \institution{AntGroup}
  \city{Hangzhou}
  \country{China}
  }
\email{jinjie.gujj@antgroup.com}

\author{Ruocheng Guo}
\authornote{Ruocheng Guo once worked at CityU and is one of Maolin Wang's supervisors. This work is not related to ByteDance.}
\affiliation{%
  \institution{ByteDance Research}
  \city{London}
  \country{UK}
  }
\email{rguo.asu@gmail.com}

\author{Xiangyu Zhao}
\authornote{Corresponding author}
\affiliation{%
  \institution{City University of Hong Kong}
  \city{Hong Kong SAR}
  \country{China}
  }
\email{xianzhao@cityu.edu.hk}
\renewcommand{\shortauthors}{Maolin Wang et al.}







\begin{abstract}
The deployment of Large Multimodal Models (LMMs) within Ant Group has significantly advanced multimodal tasks in payment, security, and advertising, notably enhancing advertisement audition tasks in Alipay.
However, the deployment of such sizable models introduces challenges, particularly in increased latency and carbon emissions, which are antithetical to the ideals of Green AI. This paper introduces a novel multi-stage compression strategy for our proprietary LLM, AntGMM. Our methodology pivots on three main aspects: employing small training sample sizes, addressing multi-level redundancy through multi-stage pruning, and introducing an advanced distillation loss design. 
In our research, we constructed a dataset, the Multimodal Advertisement Audition Dataset (MAAD), from real-world scenarios within Alipay, and conducted experiments to validate the reliability of our proposed strategy. Furthermore,
the effectiveness of our strategy is evident in its operational success in Alipay's real-world multimodal advertisement audition for three months from September 2023. Notably, our approach achieved a substantial reduction in latency, decreasing it from 700ms to 90ms, while maintaining online performance with only a slight performance decrease. Moreover, our compressed model is estimated to reduce electricity consumption by approximately 75 million kWh annually compared to the direct deployment of AntGMM, demonstrating our commitment to green AI initiatives.
\end{abstract}


\newcommand{\adr}[1]{{\color{red} [Address: #1]}}
\newcommand{\zxy}[1]{{\color{blue} [Xiangyu: #1]}}
\newcommand{\rc}[1]{{\color{brown} [Ruocheng: #1]}}
\newcommand{\zc}[1]{{\color{blue} [cite]}}





\maketitle

\input{Content/Introduction}
\input{Content/Preliminary}

\input{Content/Methodology}

\input{Content/Experiments}

\input{Content/Related_works}

\section{Conclusion and Discussion}
AntGMM's integration into Ant Group's ecosystem has significantly demonstrated the power of Large Multimodal Models in enhancing task accuracy across various sectors. However, its initial rollout brought to light two major issues: increased latency and high energy consumption, highlighting the urgent need for greener AI solutions.
In response, we developed a nuanced, multi-stage model compression strategy that efficiently mitigates these concerns, balancing high performance with environmental responsibility. This approach, specially designed for generative LMMs, has stood the test of time, maintaining exemplary performance metrics while cutting down on latency and energy costs.

\begin{acks}
This research was supported by Ant Group (CCF-Ant Research Fund, Ant Group Research Fund). This research was also partially supported by Research Impact Fund (No.R1015-23), APRC - CityU New Research Initiatives (No.9610565, Start-up Grant for New Faculty of CityU), CityU - HKIDS Early Career Research Grant (No.9360163), Hong Kong ITC Innovation and Technology Fund Midstream Research Programme for Universities Project (No.ITS /034/22MS), Hong Kong Environmental and Conservation Fund (No. 88/2022), and SIRG - CityU Strategic Interdisciplinary Research Grant (No.7020046, No.7020074).
\end{acks}

\bibliographystyle{ACM-Reference-Format}
\balance
\bibliography{llm}

\appendix
\input{Content/Appendix}

\end{document}

%% file: Content/Introduction.tex
\section{Introduction}
The advent of Large Language Models (LLMs)~\cite{brown2020language,touvron2023llama,yang2023baichuan,bai2023qwen} marks a significant milestone in Artificial Intelligence (AI). 
An even broader paradigm, the Large Multimodal Model (LMM)~\cite{zeng2023clip2,li2023blip,radford2021learning}, expands the capabilities of LLMs {by incorporating visual signals}.
LMMs excel not only in handling and generating substantial textual-only tasks, but also demonstrate impressive performance in various multimodal tasks, such as video recommendations~\cite{yi2023large}, image understanding~\cite{an2023openleaf}, and dialogue systems~\cite{delsoz2023use}. 
Particularly within our AntGroup, a leading technology company specializing in financial services and digital payments,
LMMs have achieved significant improvements in multimodal tasks related to video-text comprehension and image-text comprehension in our platform and advertising sectors.
Specifically, in the scenarios of short video categorization on Alipay, a platform renowned in China for its extensive financial services and digital payment solutions, we observed a notable $4.2\%$ enhancement in the accuracy of primary category classification.
Regarding the multimodal advertisement audition task on the same platform, we recorded a substantial $18.7\%$ increase in attribute recognition accuracy.
Furthermore, LMMs have demonstrated a significant improvement in digital advertising effectiveness, evidenced by a $1.4\%$ rise in Click-Through Rate (CTR) and a remarkable $34\%$ uplift in Per-View Click-Through Rate (PVCTR) in the context of video topic voting scenarios.

Despite the remarkable capabilities of LMMs, their deployment in particular business frameworks, such as AntGroup's multimodal advertisement audition tasks, poses significant challenges. 
Predominantly, the extensive size of these models necessitates substantial computational and storage resources~\cite{zhu2023survey}, consequently leading to increased latency—from approximately 85ms to 700ms—in the context of AntGroup's multimodal advertisement audition tasks, thereby considerably deteriorating user experience. 
Moreover, the direct implementation of LMM is associated with significant electricity consumption, a pressing concern in the quest for environmentally sustainable AI solutions. This quest aligns with AntGroup's objectives of advancing Green AI~\cite{zhou2023opportunities}, a movement aimed at fostering ecological integrity and social justice throughout the AI product lifecycle. 
A critical challenge in addressing these aspects lies in striking a balance between maintaining model performance and reducing energy and time consumption~\cite{zhou2023opportunities}. 
Model compression~\cite{zhu2023survey} emerges as a promising avenue to ameliorate LMM's inference efficiency and curtail electricity consumption without substantial compromises, thereby paving the way for more sustainable and user-centric AI solutions.

In recent years, the academic discourse has illuminated several techniques to tackle the model compression issues, centralizing on methodologies such as model pruning~\cite{gordon2020compressing,liu2021ebert}, knowledge distillation~\cite{sun2019patient,jiao2020tinybert}, and quantization~\cite{kim2021bert,shen2020q}. Nevertheless, the majority of the existing literature~\cite{xu2023survey} on transformer model compression is predominantly anchored in the BERT-style pre-training and fine-tuning paradigm. These compression strategies are primarily tailored for natural language understanding (NLU) tasks and exhibit a lack of direct applicability to the contemporary generative LMMs~\cite{zhu2023survey}.
For compressing generative decoder-only LLMs, we believe there are the following distinct challenges:

\begin{enumerate}[leftmargin=*]
    \item \textbf{Challenge 1: Retraining LLMs is Cost-prohibitive.} 
    Many earlier compression techniques depended on extensive retraining over large corpora~\cite{xu2023survey}. However, as LLMs or LMMs~\cite{brown2020language,touvron2023llama,yang2023baichuan,bai2023qwen} have scaled the corpus size to trillions of tokens or more, retraining has become infeasible.
    \item\textbf{Challenge 2: Identifying Multi-level Redundancy is Difficulty.} 
    Boasting a parameter count of at least 1 billion or more alongside an intricate structure, the redundancy within LLMs or LMMs is apparent at multiple levels, encompassing \textbf{the number of blocks}, \textbf{input dimensions}, and \textbf{intermediate-module dimensions}. Addressing redundancy across these levels in one shot will result in a non-smooth impact on performance~\cite{syed2023prune}.
    \item \textbf{Challenge 3: Sequential Token Generation Complicate Compression Task.}
     Generating extended texts is inherently challenging due to the cascading nature of the task, where each step entails large-scale classification (e.g., a vocabulary size of over 100,000 implies more than 100,000 categories). A divergence from the ground truth for a single token during generation can cause subsequent tokens to deviate considerably. It's worth noting that the loss during the distillation phase requires the specialized design to address these challenges effectively.
\end{enumerate}

Despite exploring unstructured pruning~\cite{frantar2023sparsegpt,sun2023simple}, structured pruning~\cite{ma2023llmpruner,fang2023depgraph}, or knowledge distillation techniques for generation tasks~\cite{shridhar2023distilling,gu2023knowledge}, existing methods fall short in addressing three main challenges. To align with Green AI goals and boost efficiency, we propose a multi-stage compression strategy for our pre-trained LMM, the Antgroup General Multimodal Model (AntGMM), targeting these challenges. Our key innovations include:

\begin{itemize}
    \item Fine-tuning on small datasets by inheriting parameters from the original network and alternating between structural pruning and knowledge distillation to minimize loss. This approach doesn't require large datasets for retraining.
    \item Addressing multi-level redundancy via multi-stage pruning, targeting redundancies in layer numbers, module hidden dimensions, and input dimensions, ensuring consistent performance and enhanced inference efficiency.
    \item Introducing a novel distillation loss for generative tasks that considers both token accuracy and contextual appropriateness, using a pairwise loss to prioritize correct responses, thereby aligning output logits more effectively.
\end{itemize}

This strategy aims to optimize and amplify the benefits of AntGMM while overcoming existing limitations.

Our framework has been operational in Alipay's multimodal advertisement audition scenario commencing from \textbf{September 2023}. 
Significantly, our framework not only maintained online performance, with only a marginal decrease of $0.8\%$ in the seven-day average accuracy, but also significantly reduced latency, cutting it down from \textbf{700ms} to \textbf{90ms}. Simultaneously, in comparison to the direct deployment of the AntGMM, our compressed model is projected to reduce electricity consumption by approximately \textbf{75 Megawatt-hour} annually.

%% file: Content/Preliminary.tex
\section{Preliminary}

\subsection{AntGMM: Ant Group's Foundational LMM}

\begin{figure}[t]
    \centering
    \includegraphics[width=\linewidth]{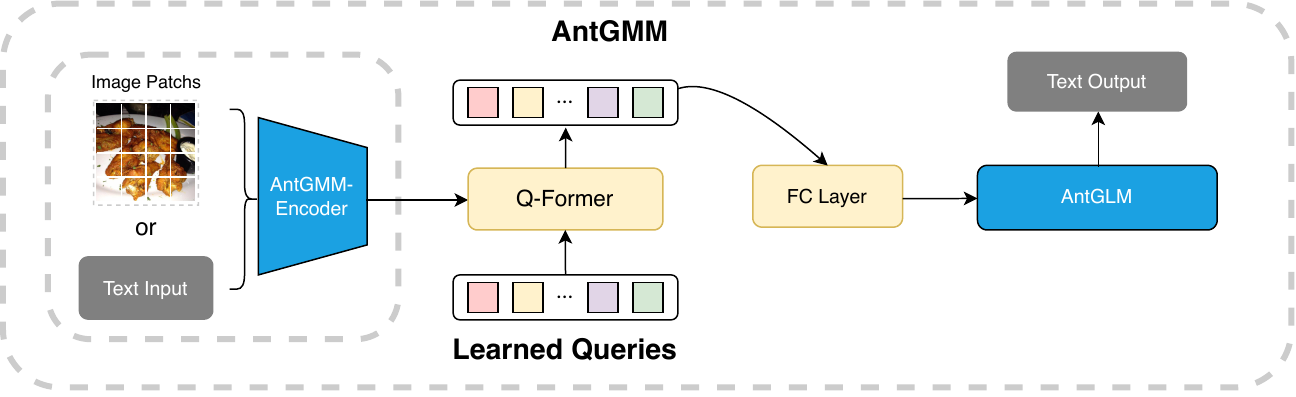} 
    \caption{Model structure of concurrent version of AntGMM. We adopted a structure akin to BLIP-2 as the cornerstone within the Ant Group. }
    \label{fig:Antgmm}
\end{figure}

At Ant Group, we have developed and pre-trained a proprietary LMM, termed Antgroup General Language Model (AntGMM). The current version of AntGMM adopts a BLIP-2-like~\cite{li2023blip} architecture that facilitates the integration of a multimodal base model, the AntGMM Encoder, with a comprehensive NLP model, termed as Antgroup General Language Model (AntGLM). As illustrated in Figure~\ref{fig:Antgmm}, AntGMM bridges the modality divide utilizing a novel Querying Transformer (Q-Former) subjected to a dual-stage pre-training process. It first partakes in vision-language representation learning using a static image encoder and subsequently transitions to a vision-to-language generative learning stage leveraging a static large language model. The AntGLM is pre-trained on a blend of public datasets and an extensive corpus of proprietary business data. To enhance business relevance, the AntGMM Encoder and the Q-Former have been pre-trained on our internal datasets.
The AntGMM is the foundation of Ant Group's multimodal business applications, marking a notable step in bringing together visual and textual data, thus enabling more intuitive and enhanced interactions in business-focused applications. It should be noted that the AntGMM architecture is subject to ongoing iterative refinements aligned with business requirements. In this paper, we will only focus on the early version of the AntGMM.

\subsection{Structured Pruning of Neural Networks}

Structured pruning~\cite{anwar2017structured,ma2023llmpruner} is a methodology that aims at mitigating the computational and memory overhead associated with neural networks, which is of paramount importance for models deployed on resource-constrained devices. Unlike unstructured pruning~\cite{kwon2020structured,frantar2023sparsegpt} that relies on sparse operations, structured pruning entails the removal of entire neurons or layers from the neural network. This leads to a model with a reduced architecture that is better suited for hardware accelerations~\cite{li2020efficient}.
The objective of structured pruning can be succinctly formulated as an optimization problem: For a neural network $f$ characterized by a weight tensor $W$, the aim of structured pruning is to identify a more compact network $g$ with a reduced weight tensor $W'$. This reduction in the network's size should incur minimal performance loss $\Delta P(\mathcal{D})$ on a dataset $\mathcal{D}$. The performance decrement is defined as the absolute difference in a performance metric $P$, such as cross-entropy, between the full and pruned networks: $\Delta P(\mathcal{D}) = |P(f,\mathcal{D}) - P(g,\mathcal{D})|$. 
To ascertain the significance of each parameter in $f$ and decide which ones to prune to get $g$, various criteria are employed by different structured pruning strategies. These often include the magnitude of weights to gauge each parameter's importance\cite{molchanov2019importance}. In this paper, we adopt a structured pruning strategy that leverages a first-order approximation to guide our pruning decisions. 

\subsection{Knowledge Distillation}

Knowledge Distillation (KD)~\cite{hinton2015distilling,shridhar2023distilling,gu2023knowledge} is a prominent model compression technique where a compact and efficient student neural network $g$ is trained to replicate the behavior of a larger, pre-trained teacher neural network $f$. In the seminal KD framework proposed by Hinton~\cite{hinton2015distilling}, the distillation procedure entails an optimization process where the student network minimizes a composite loss function. This function is a combination of a ground truth loss $\mathcal{L}_{CE}(g(x),y)$ and a fidelity loss $L_{KD}(g(x), f(x))$, where $x$ denotes the input data and $y$ represents the corresponding ground truth labels. However, further investigations in the realm of Natural Language Processing (NLP) have demonstrated that exclusively focusing on soft aligning via fidelity loss $\mathcal{L}_{KD}(g(x), f(x))$ {can enhance the student model's performance}~\cite{zhang2020distilling,luo2020knowledge}.
For straightforward classification tasks, this distillation loss can be operationalized by aligning the logit outputs of the teacher and student models through the Kullback-Leibler (KL) divergence~\cite{shlens2014notes}. Nonetheless, when addressing the intricacies of generative tasks, reliance on KL divergence alone proves to be inadequate for LMMs~\cite{gu2023knowledge}. We designed a pair-wise loss approach to recover performance.

%% file: Content/Methodology.tex
\section{Methodology}

In this section, we will present our compression algorithm in detail. The parameter notations of this paper are detailed in Table~\ref{tab:notation}.

\begin{table}[t]
\centering
\caption{Notations used in our algorithm}
\resizebox{\linewidth}{!}{
\begin{tabular}{cl}
\hline
\textbf{Symbol} & \textbf{Description} \\
\hline
$f$ & LMM model \\
$g$ & Pruned LMM model \\
$\mathcal{D}_p$ & Pruning dataset \\
$\mathcal{D}_d$ & Distillation dataset \\
$\mathcal{D}_t$ & Test dataset \\
$\alpha$ & Maximum performance tolerance \\
$E$ & Number of post-distillation epochs \\
$N_{blocks}$ & Maximum number of pruned blocks \\
$n_{b}$ & Number of blocks pruned at each step\\
$N_{inter}$ & Maximum reduction in Inter-Module Dimensions \\
$n_h$ & Number of Inter-Module Dimensions pruned at each step \\
$N_{input}$ & Maximum reduction in input dimensions \\
$n_d$ & Number of input dimensions pruned at each step \\
\hline
\end{tabular}}
\label{tab:notation}
\end{table}

\begin{algorithm}[t]
\caption{Language Multimodal Model Pruning Procedure}
\label{alg:overview}

\DontPrintSemicolon
\KwIn{$f, \mathcal{D}_p, \mathcal{D}_d, \mathcal{D}_t, \alpha, E, N_{blocks}, N^{(Att)}_{inter},N^{(FFN)}_{inter}$,\\$N_{input}, d, n_{b}, n^{(FFN)}_{h},n^{(Att)}_{h},n_d$}
\KwOut{Pruned model $g$}

Calculate performance $p_f$ of $f$ using $\mathcal{D}_d$\;
Initialize pruned LMM $g_0 \gets f$\;

\tcc{Stage 1: Block Pruning}
\While{$p_f-p_{g_0}  < \alpha$ \textbf{and} Not reached maximum number of pruned blocks $N_{blocks}$}{
    Remove the last $n_{b}$ block from the largest module in $g_0$\;
    Distill $g_0$ for $E$ epochs using loss from Eq.~\ref{eq:distill} and $\mathcal{D}_d$\;
    Calculate current performance $p_{g_0}$ with $\mathcal{D}_d$\;
}
$g_1 \gets g_0$\;

\tcc{Stage 2.1: Inter-Module Pruning in FFN}
\While{$p_{g_0}-p_{g_1}  < \alpha$ \textbf{and} Not reached maximum Inter-Module Dimension reduction $N^{(FFN)}_{inter}$}{
    \For{each block in $g_1$}{
        Prune $n^{(FFN)}_{h}$ hidden dimensions of FFN via Eq.~\ref{eq:FFN}\;
    }
    Distill $g_1$ for $E$ epochs via Eq.~\ref{eq:distill} and $\mathcal{D}_d$\;
    Calculate current performance $p_{g_1}$ with $\mathcal{D}_d$\;
}
$g_2 \gets g_1$\;

\tcc{Stage 2.2: Inter-Module Pruning in Att}
\While{$p_{g_1}-p_{g_2}  < \alpha$ \textbf{and} Not reached maximum Inter-Module Dimension reduction $N^{(Att)}_{inter}$}{
    \For{each block in $g_2$}{
        Prune $n^{(Att)}_{h}$ hidden dimensions in Att via Eq.~\ref{eq:Att}\;
    }
    Distill $g_2$ for $E$ epochs using loss from Eq.~\ref{eq:distill} and $\mathcal{D}_d$\;
    Calculate current performance $p_{g_2}$ with $\mathcal{D}_d$\;
}
$g_3 \gets g_2$\;

\tcc{Stage 3: Input/Output Dimension Pruning}
\While{$p_{g_2}-p_{g_3}  < \alpha$ \textbf{and} Not reached maximum input-dimension reduction $N_{input}$}{
    Prune $n_d$ input/out dimensions using Eq.~\ref{eq:inout}\;
    Distill $g_3$ for $E$ epochs using loss from Eq.~\ref{eq:distill} and $\mathcal{D}_d$\;
    Calculate current performance $p_{g_3}$ with $\mathcal{D}_d$\;
}
\Return $g \gets g_3$\;
\end{algorithm}

\subsection{Overview}

This section outlines a methodical strategy for effectively minimizing an LMM's size while maintaining its performance, described through a three-phase process as shown in Algorithm~\ref{alg:overview} for structured downsizing. The \textit{Block Pruning} phase initially reduces the LMM's block numbers by pruning several of the last blocks. Following this, the \textit{Inter-Module Dimension Pruning} phase refines the structure by targeting the Feedforward Networks (FFNs) and Attention mechanisms within the LMM, involving a granular optimization where dimensions within blocks are assessed and pruned. The final \textit{Input-Dimension Pruning} stage adjusts the input sizes for each block to match the pruned structure, thus allowing the LMM to operate more efficiently without the unnecessary computational costs of processing oversized inputs. Throughout the pruning stages, the model is subjected to performance thresholds to prevent significant degradation of the model's accuracy, with every pruning process interspersed with several distillation epochs to compensate for any potential loss in performance due to the reduction in model complexity.
This approach promises a balance between model efficiency and performance, making deploying LMMs more practical in resource-constrained environments.

\subsection{Element-wise Importance Caculation}

In the realm of neural network pruning, traditional metrics such as the L1 norm ~\cite{zafrir2021prune} or magnitude-based approaches~\cite{lee2020layer} are foundational. They provide a direct and sensitive means to assess the impact of individual weights on network performance. Within this domain, first-order and second-order techniques are of particular significance.
First-order~\cite{hou2020dynabert} gradients are of considerable importance in this context. They offer insightful reflections on the influence of weight modifications on the loss function. This is especially relevant in scenarios with limited data availability, where a direct evaluation of each weight’s contribution is essential~\cite{wang2019eigendamage}.

Motivated by these challenges, we have utilized first-order gradient information to estimate the importance of weights. Suppose given a small dataset $\mathcal{D}=\{X,Y\}$, the importance of one specific parameter $\theta$ in the neural network $ f$ can be formulated as:
\begin{align}
    I_{\theta}=|\Delta \mathcal{P}(\mathcal{D})|=\left|\mathcal{P}_{f}(\mathcal{D})-\mathcal{P}_{f_{\theta=0}}(\mathcal{D})\right| \approx
    \left|\frac{\partial \mathcal{P}_{f}(\mathcal{D})}{\partial \theta} \theta+\mathcal{O}\left(|\theta|^2\right)\right|
    \label{eq:1}
\end{align}
where $f_{\theta=0}$ represents the neural network in which the specific parameter $\theta$ has been pruned,  i.e., assigned $\theta$ a value of zero. Evaluation of $P$ can be articulated through the loss function $\mathcal{L}$, the deviation $|\frac{\partial\mathcal{L}(\mathcal{D})}{\partial\theta}\theta|$ can serve as a proxy for assessing the importance of the single parameter $\theta$. 

\begin{figure}[t]
    \centering
    \includegraphics[width=\linewidth]{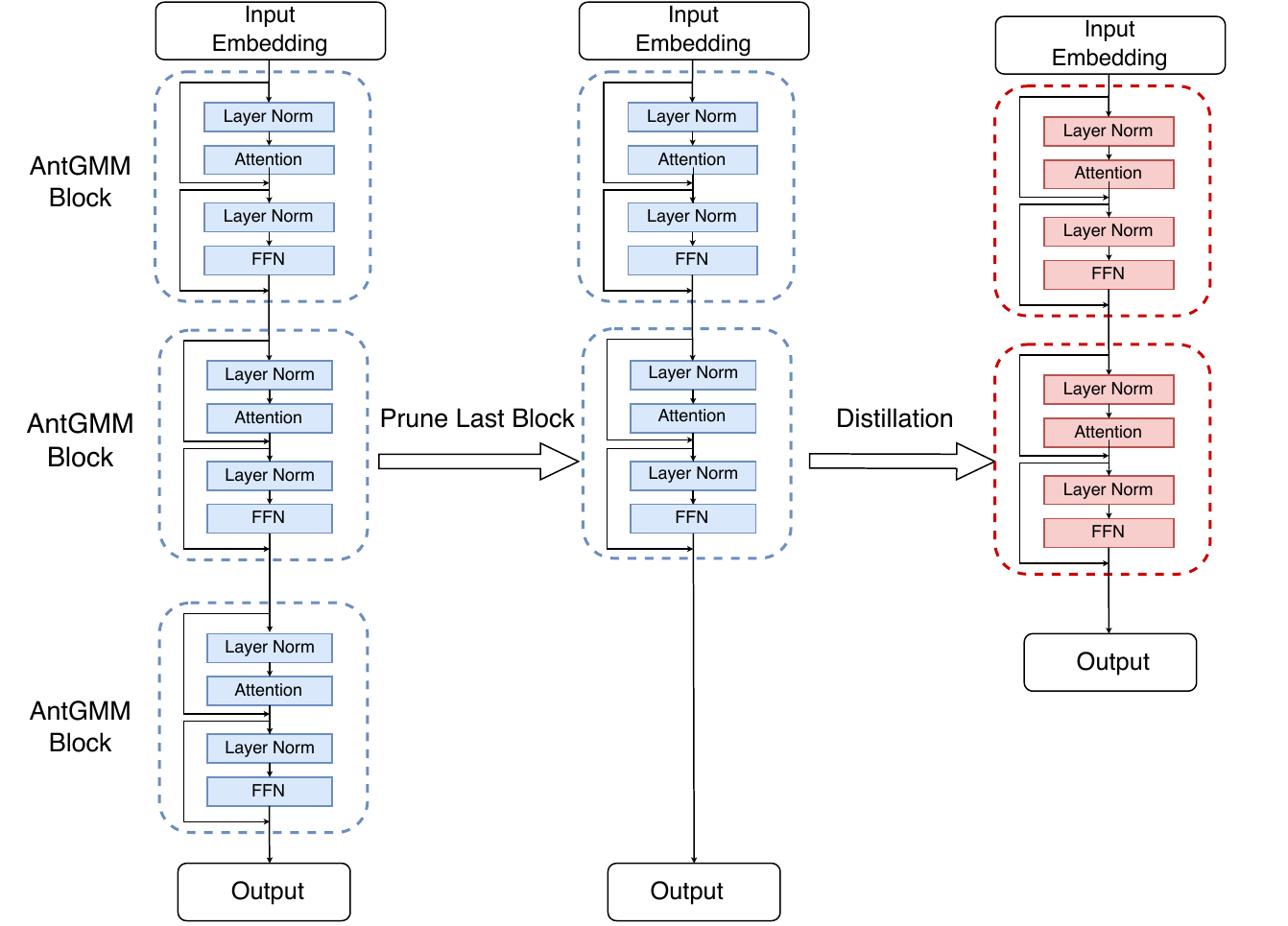} 
    \caption{Block pruning. Block pruning involves removing the last layer (or a few last layers) at each iteration, followed by distillation to ensure minimal loss in model performance.}
    \label{fig:layer}
\end{figure}

\subsection{Multi-stage Pruning}
Facing the intricate challenge of multi-level redundancy in large models with billions of parameters, our research introduces a multi-stage pruning approach to systematically enhance efficiency while maintaining model performance.
\subsubsection{Stage 1: Block Pruning}
As depicted in Figure~\ref{fig:layer}, our methodology initiates by pruning blocks to reduce model complexity, thereby decreasing its depth. Our process begins with removing the final block, followed by iterative model distillation until convergence is achieved. This step is repeated—each time removing an additional block—until a significant increase in loss is detected. 
The total parameter count of the models is directly tied to the remaining number of blocks; hence, directly reducing blocks results in a substantial decrease in model complexity. However, this is the coarsest level of granularity and has a considerable impact on the effectiveness of the model. 

\begin{figure}[t]
    \centering
    \includegraphics[width=1\linewidth]{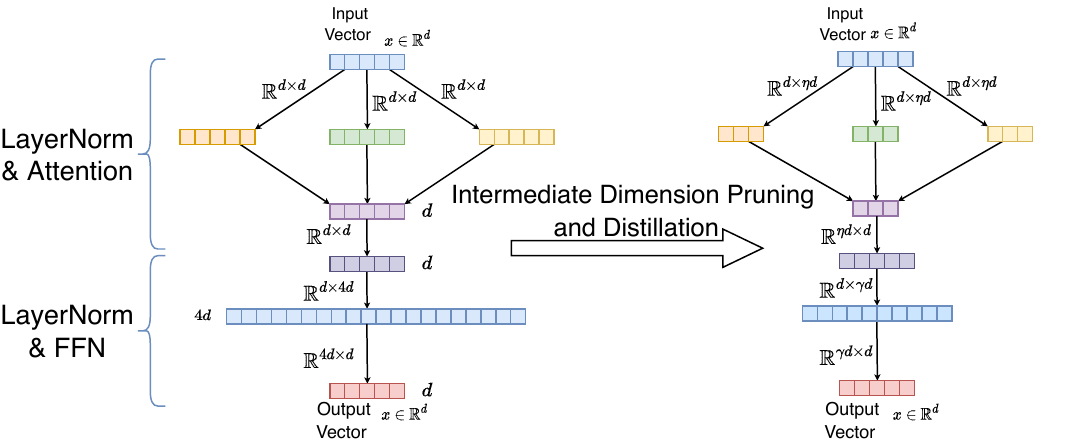} 
    \caption{{Intermediate-module dimention pruning. The second stage focuses on dimensionality pruning of the hidden layers of FFNs and attention mechanisms within blocks.}}
    \label{fig:inter}
\end{figure}
\begin{figure}[t]
    \centering
    \includegraphics[width=1\linewidth]{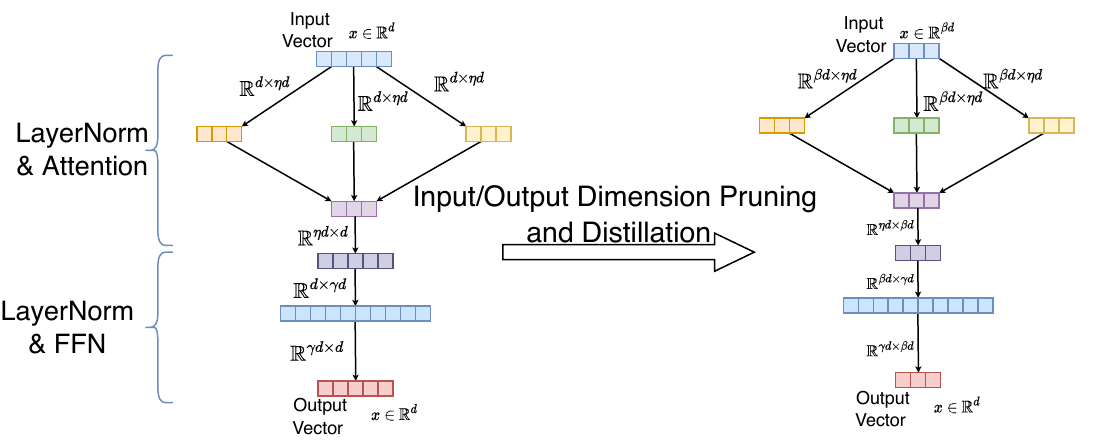} 
    \caption{{Input dimention pruning. The final stage reduces the number of parameters associated with both the input and output dimensions. This requires synchronized reduction across all AntGMM blocks.}}
    \label{fig:input}
\end{figure}
\subsubsection{Stage 2: Inter-Module Dimension Pruning}

In the second stage of our process, as shown in Figure~\ref{fig:inter}, we concentrate on dimensionality reduction of the hidden layers via a combination of pruning and knowledge distillation. This stage focuses on the Inter-Module Dimensions of feed-forward networks (FFNs) and attention mechanisms within AntGMM blocks, identifying and pruning redundant parameters based on their calculated importance.

To ascertain the significance of each Inter-Module Dimension, which is associated with many parameters, we aggregate the importance by summation. The importance of the $n$-th Inter-Module Dimension in an FFN, $y = \text{FFN}(x) = (\mathbf{W}^{(\text{out})} \text{GELU}(\mathbf{W}^{(\text{in})} x))$, with parameters $\mathbf{W}^{(\text{out})} \in \mathbb{R}^{4d \times d}$ and $\mathbf{W}^{(\text{in})} \in \mathbb{R}^{d \times 4d}$, is computed as:
\begin{equation}
    I_{\text{inter}}^{(\text{FFN})}(n) = \sum_{i=1}^{d} I(\mathbf{W}_{in}^{(\text{in})}) + \sum_{j=1}^{4d} I(\mathbf{W}_{nj}^{(\text{out})}),
\label{eq:FFN}
\end{equation}
where $I(\cdot)$ denotes the importance measure as defined in Eq.~\ref{eq:1}, $d$ is the dimension number of input vectors.

Regarding the attention mechanism, the importance of the $n$-th dimension in $h$-th head is calculated using the significance of the attention operation parameters $\mathbf{W}_{h}^{(Q)}$, $\mathbf{W}_{h}^{(K)}$, and $\mathbf{W}_{h}^{(V)}$, along with the attention projecting parameters $\mathbf{W}_{h}^{(O)}$, as follows:
\begin{align}
I_{\text{inter}}^{(\text{Att})}(n) &= \sum_{h}\sum_{i=1}^{d} \left(I(\mathbf{W}_{hin}^{(Q)}) + \sum_{j=1}^{d} I(\mathbf{W}_{hin}^{(K)}) 
\right. \nonumber \\
&\quad + \left.
\sum_{k=1}^{d} I(\mathbf{W}_{hin}^{(V)}) + \sum_{l=1}^{d} I(\mathbf{W}_{hnl}^{(O)})\right).
\label{eq:Att}
\end{align}
The pruning of each block is conducted independently, yet for the sake of uniformity, we ensure that the dimension pruned in each block is consistent in every iteration.
Pruning and distillation are performed iteratively until a significant loss in performance is noted. This results in reduced dimensions of the weight matrix, particularly adjacent to the output layer. As illustrated in Figure~\ref{fig:inter}, this stage effects a substantial reduction in the parameters.

\subsubsection{Stage 3: Input/Output Dimension Pruning}
In the final stage, as shown in Figure~\ref{fig:input}, we concentrate on reducing the number of parameters associated with input and output dimensions, we concentrate on reducing the number of parameters associated with input and output dimensions. Unlike Inter-Module Dimension pruning, pruning the input/output dimension necessitates synchronized reduction across all blocks and associated modules (this includes the embedding module, layer normalization, attention, and FFNs) to ensure dimension consistency. Specifically, for the $n$-th input/output dimension, its importance is defined as:
\begin{align}
I^{(\text{In/Out})}(n) &= \sum_b\left(\sum_{i} I(\mathbf{W}_{ni}^{(b)(\text{in})}) + \sum_{j} I(\mathbf{W}_{jn}^{(b)(\text{out})}) \right. \nonumber \\
&\quad + \left. \sum_{h}\left( \sum_{k} I(\mathbf{W}_{hnk}^{(b)(Q)}) + \sum_{l} I(\mathbf{W}_{hnl}^{(b)(K)})  \right.\right. \nonumber \\
&\quad 
\left. \left. 
+\sum_{p} I(\mathbf{W}_{hnp}^{(b)(V)}) + \sum_{q} I(\mathbf{W}_{hqn}^{(b)(O)})\right)\right),
\label{eq:inout}
\end{align}

where $\mathbf{W}^{(b)}$ represents the parameters within the $b$-th block.
It is noteworthy that parameters across all blocks are pruned. Hence, this stage culminates in a reduction in the size of all weight matrices, significantly compressing the model. As depicted in Figure~\ref{fig:input}, this phase achieves a further decrement in the number of parameters.
\subsection{Distillation Loss Design}


Addressing the complexity of sequential token generation in text synthesis, where even a single token deviation can cascade into significant errors, we innovate on distillation loss design to optimize compression in large-scale language models.

Specifically, in the distillation phase for generative models, we address the complexity of sequential token generation by introducing a pairwise loss function. This key element of our approach ensures prioritization of the logits for correct token responses over those for incorrect tokens, thereby explicitly enhancing the model's ability to generate contextually accurate responses.
The distillation loss function is defined as:
\begin{align}
      \mathcal{L}_{distill} &= \sum_{i=1}^{\text{Length}}\left( \text{KL}(g_i(x), f_i(x)) + \right. 
      \nonumber \\
&\quad 
\left.
      \gamma \max(0, \mathbf{Prob}^{i}_{f}(\text{Token}=T^{g}_{i}) - \mathbf{Prob}^{i}_{g}(\text{Token}=T^{g}_{i}))\right)  
    \label{eq:distill}
\end{align}
where $\text{Length}$ signifies the maximum token length in the sequence, $T^{g}_{i}$ is the token with the highest probility for $g$ at the $i$th position, $\gamma$ is a hyperparameter that balances the pairwise loss, and $\mathbf{Prob}_{g}$ denotes the probability that network $g$ assigns to a specific token at the $i$th position. This function aims to minimize the KL divergence between the probability distributions of the teacher and student networks while ensuring that the student's confidence in predicting the actual tokens is at least on par with the teacher's.
The pairwise loss element in our approach addresses the tendency of the model to make frequent but incorrect predictions by adjusting the logits. This ensures that the logits for the correct answers could surpass those for the most incorrect answers. Implementing this mechanism effectively lowers the chance of consistent incorrect responses, fostering a more precise and appropriate generation.

%% file: Content/Experiments.tex
\section{Experiments}

In the context of multimodal scenarios, pre-training of image semantics is a prerequisite. However, for the specific task of multimodal advertisement auditing, we face a unique challenge: the absence of both \textbf{a dedicated open dataset} and \textbf{a baseline model with appropriately aligned semantics}. To overcome this, we constructed a multimodal advertisement auditing dataset from our own business and exclusively employed our pre-trained model, AntGMM. This model was specifically developed to meet the unique requirements of this task.
In this section, we will conduct experiments on our ad-auditing dataset to answer the following research questions.
\begin{itemize}[leftmargin=*]
\item \textbf{RQ1}: What is the performance at each stage?
\item \textbf{RQ2}: Can our approach maintain efficacy when applied to LMMs with a small amount of data? (Challenge 1)
\item \textbf{RQ3}: Does minimizing redundancy in one stage suffice for satisfactory performance? (Challenge 2)
\item \textbf{RQ4}: How does the performance compare between iterative pruning and one-shot pruning? (Challenge 2)
\item \textbf{RQ5}: In our specific task, does employing hard label distillation adversely affect performance? (Challenge 3)
\item \textbf{RQ6}: What is the impact of integrating pairwise loss on the model's overall performance? (Challenge 3)

\end{itemize}
\subsection{Datasets} 
Initially, we sought a suitable public dataset but were unable to locate any that were pertinent to our specific requirements. To advance real-world deployment, we extracted some real-world advertising data from Alipay and compiled a dataset, referred to as the Multimodal Advertisement Audition Dataset (MAAD), designed to enable the pruning and distillation of AntGMM. We are currently anonymizing user information within MAAD to safeguard user privacy, and we are also undergoing a legal review to ensure compliance with relevant regulations. The release of the MAAD dataset is scheduled for about ~\textbf{February}, with the aim of contributing to the wider research community's efforts in LMMs' applications.
\begin{figure}[t]
    \centering
    \includegraphics[width=0.8\linewidth]{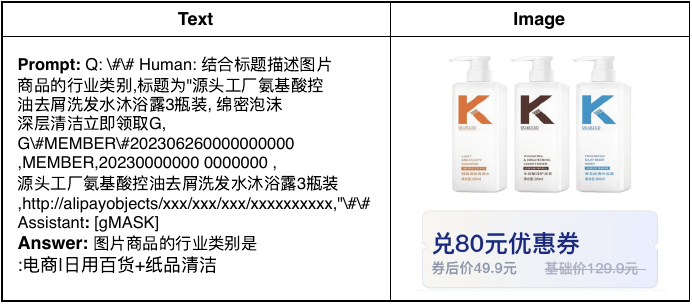} 
    \caption{A data sample from the dataset. It involves an advertisement image accompanied by a Chinese text segment. The text includes a carefully constructed prompt, possible assistant details, and a classification description of the advertisement. Some sensitive information has been processed.}
    \label{fig:dataset}
\end{figure}
As shown in Figure~\ref{fig:dataset}, for multimodal audition scenarios, each data entry in our dataset comprises an image, a question, and an answer. 
In our study, we meticulously crafted prompts to leverage large models for our specific scenario. Our dataset encompasses 12,000 training samples and 3,000 test samples. 
Specifically, our data is composed of five distinct tasks: recognition of advertisement background color, identification of manifestations, classification of specific advertisement content, recognition of advertisement logos, and identification of advertisement styles. Each task is designed to probe a different aspect of the model's capability to understand and process visual information pertaining to advertisements.
This paper evaluated all tasks using ~\textbf{classification accuracy} as the metric. 

\subsection{Implementation Details} 
For the compression of AntGMM, a single process can be conducted on one A100 GPU. All our experiments were carried out on a machine equipped with eight A100s. Our inference time was assessed on a single A10 GPU.
We employed the AdaW optimizer with a learning rate of 3e-5. Drawing inspiration from the fine-tuning approach of LLaMA, we implemented a gradient schedule strategy using the cosine function. For the importance calculation during pruning and the knowledge distillation, we both utilized the entire training set of 12k samples. Our batch size was set to 1024.
Regarding the parameters in our algorithm, we reduced the number of blocks by $n_{b} = 4$ at each pruning iteration, with a maximum reduction of $N_{blocks} = 40$ blocks. The intermediate-module layer dimension of the FFN was reduced by $n^{(FFN)}_h = 768$ at each iteration, with a maximum reduction of $N^{(FFN)}_{inter} = 6144$. The intermediate-module layer dimension of attention was reduced by $n^{(Att)}_h = 96$ at each iteration, with a maximum reduction of $N^{(Att)}_{inter}=768$. The input/output dimensions were reduced by $n_d = 96$ at each iteration, with a maximum reduction of $N_{input} = 768$. We set the maximum tolerance for the effect degradation to $\alpha=0.15$. After each pruning iteration, we conducted distillation for $E = 4$ epochs.
\begin{figure}[t]
    \centering
    \includegraphics[width=0.85\linewidth]{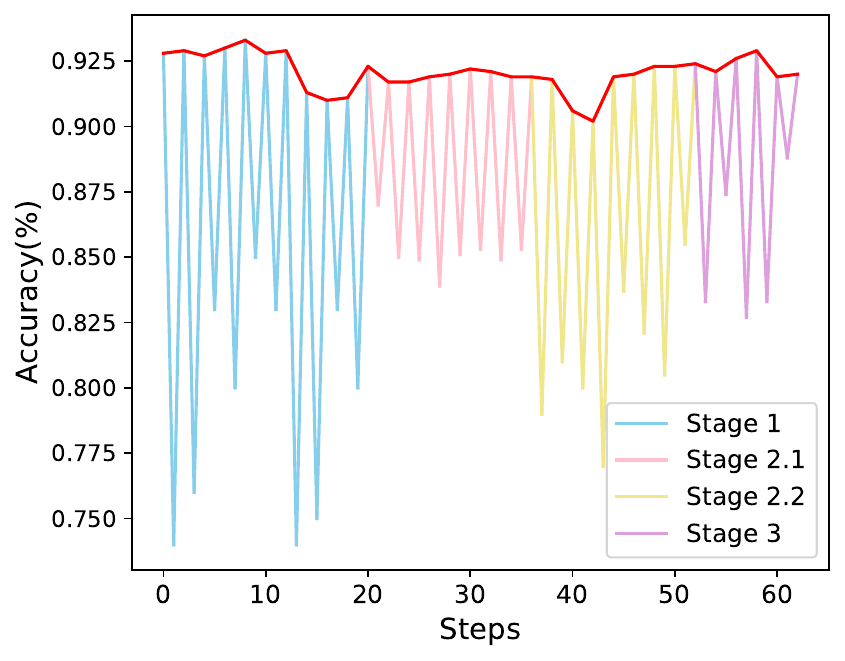} 
    \caption{We present the results of each step, with a particular emphasis on the evaluation that pruning and distillation were \textbf{separated into two steps} in this figure. After each pruning step, which reduces performance, a distillation step follows to recover and enhance the performance throughout the process. The red line represents the overall performance variations where pruning and distillation are \textbf{combined as one step} throughout the entire process. }
    \label{fig:distillation_per}
    \vspace{-0.5cm}
\end{figure}

\subsection{\textbf{RQ1}: Overall Performance }
\begin{table*}[ht]
\centering
\caption{Overall Model Performance}
\label{tab:model_metrics}
\resizebox{\textwidth}{!}{
\begin{tabular}{
  l
  S[table-format=1.2]
  S[table-format=1.2]
  S[table-format=1.3]
  l
  l
  l
  l
  S[table-format=1.3]
}
\toprule
\thead{Model} & {\thead{Size\\(B)}} & {\thead{Inference Time on A10\\(samples/s)}} & {\thead{Color}} & {\thead{Manifestations}} & {\thead{ Content}} & {\thead{Logo}} & {\thead{Style}} & {\thead{Overall Accuracy}} \\
\midrule
\makecell{Original AntGMM} & 5.4 & 1.38 & 0.932 & 0.967 & 0.934 & 0.975 & 0.833 & 0.928 \\
\makecell{After Block Pruning} & 0.9 & {6.01}& 0.923 & 0.951 & 0.925 & 0.966 & 0.802 & 0.914 \\
\makecell{After Inter-Module Pruning} & 0.6 & {6.70} & 0.924 & 0.968 & 0.931 & 0.977 & 0.806 & 0.920 \\
\makecell{After Input/Output Pruning} & 0.3 & {7.62} & 0.920 & 0.971 & 0.916 & 0.981 & 0.825 & 0.923 \\
\bottomrule
\end{tabular}}
\end{table*}
In this study, we applied our optimization framework to the AntGMM model, achieving substantial improvements. As detailed in Table~\ref{tab:model_metrics}, our method markedly reduced the model's parameter count by approximately 18 times, from 5.4 billion to 0.3 billion. This reduction was accompanied by a significant increase in inference efficiency on the A10 processor, with a performance boost from 1.38 to 7.62 samples per second, which constitutes an enhancement of roughly 7.7 times.
Despite this considerable decrease in parameters, the model's performance on the test dataset remained robust without a significant decline. Notably, task-specific performance showed an uptick, with accuracy in Manifestations rising from 0.967 to 0.971 and in Logo recognition from 0.975 to 0.981. Block pruning played a pivotal role in these improvements, significantly reducing the parameter count to 0.9 billion and increasing inference efficiency to 6.01 samples per second.

Figure~\ref{fig:distillation_per} illustrates the critical role of distillation in recuperating performance post-pruning. Following each pruning phase, which initially diminishes performance, a subsequent distillation phase is employed to recover and even augment performance.
The pruning of intermediate-module dimensions in the FFN and the input/output dimensionality had a smaller impact on performance, as there is less fluctuation in these stages compared to others. In the other two stages, performance was still effectively mitigated through distillation.
Our method not only significantly reduced the number of parameters in the model and improved inference efficiency but also maintained model performance across most tasks and even enhanced it in specific tasks.

\begin{figure}[t]
    \centering
    \includegraphics[width=0.7\linewidth]{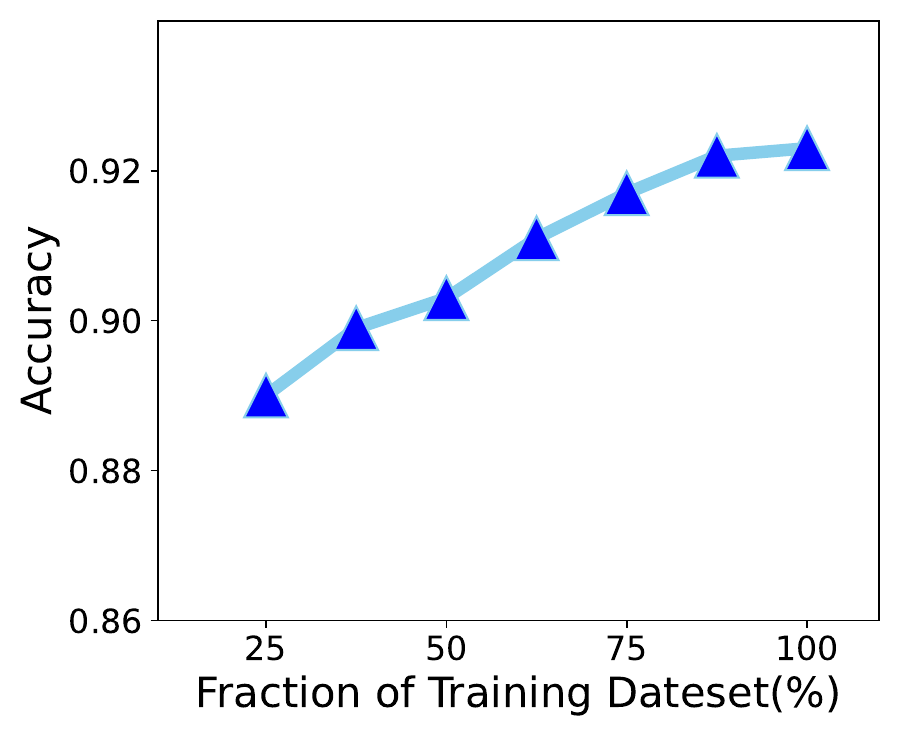} 
    \caption{Correlation between sample size and performance.}
    \label{fig:sample_performance}
\end{figure}
\subsection{\textbf{RQ2}: Training Samples and Performance }

We also conducted a comprehensive analysis of the correlation between the final performance and the distillation sample size. The results depicted in the illustrated Figure~\ref{fig:sample_performance} demonstrate a strong positive correlation between the sample size and the final compression effect of the model. Interestingly, it was observed that even with a sample size as small as 3k, desirable outcomes can be achieved. However, it is worth noting that a sample size of 12k proves to be sufficient in yielding results comparable to the original AntGMM. Consequently, the model that was eventually deployed online employed a sample size of 12k.

\subsection{\textbf{RQ3}: Only One Level Pruning }

\begin{table}[t]
\centering
\caption{Comparison of Pruning Methods}
\label{tab:model_performance:2}
\resizebox{\linewidth}{!}{
\begin{tabular}{
  l
  S[table-format=1.1]
  S[table-format=2.2]
  S[table-format=1.3]
}
\toprule
{Model} & {\thead{Model Size\\ (B)}} & {\thead{Inference Time in A10\\ (samples/s)}} & {Accuracy} \\
\midrule
Block Pruning Only                                  & 0.7 & 7.93 & 0.903 \\
Inter-Module(FFN) Pruning Only                  & 2.2 & 2.11 & 0.785 \\
Inter-Module(Att) Pruning Only            & 3.8 & 1.87 & 0.080 \\
Input/Output Pruning Only                               & 0.6 & 1.95 & 0.061 \\
Block Pruning Only (One-shot)                       & 0.7 & 7.93 & 0.863 \\
Inter-Module(FFN) Pruning Only (One-shot)       & 2.2 & 2.11 & 0.530 \\
Inter-Module(Att) Pruning Only (One-shot) & 3.8 & 1.87 & 0.076 \\
Input/Output Pruning Only (One-shot)                    & 0.6 & 1.95 & 0.033 \\
All Stages (One-shot)                                     & 0.3 & 7.62 & 0.021 \\
All Stages                                         & 0.3 & 7.62 & 0.923 \\
\bottomrule
\end{tabular}}
\end{table}

We also examined a single-stage pruning method, focusing on reducing individual redundancies rather than a collective approach to model compression. Referencing Table~\ref{tab:model_performance:2}, pruning solely within the FFN of the intermediate module resulted in a significant drop in accuracy to 0.785. The effect was even more dramatic when pruning targeted only the intermediate-module dimensions of attention layers or the input/output dimensions, leading to a steep decline in performance. In particular, pruning just the input/output dimensions led to an accuracy of merely 0.061. These findings emphasize the high sensitivity of both the intermediate-module of attention and input/output dimensions to pruning. Therefore, our results suggest that pruning at a single level is inadequate for preserving an ideal balance between model size and performance during LMM compression, highlighting the limitations of single-stage pruning.

\subsection{\textbf{RQ4}: One-shot Pruning vs. Gradual Pruning}

In addition, we also investigated the one-shot pruning approach, aiming for immediate attainment of the desired model structure followed by distillation. The findings highlighted the essential nature of our step-by-step pruning method. As shown in Table~\ref{tab:model_performance:2}, one-shot pruning significantly impaired model performance to the degree that subsequent distillation could not recover. There was apparent in the sharp decline in accuracy, for example, from 0.785 to 0.530 in the one-time pruning of the FFN. 
Interestingly, the one-shot pruning of dimensions within the intermediate-module dimension of the attention or the input/output dimension proved the most harmful to performance. This is consistent with the findings of the only one-level pruning experiments. These results underscore the importance of careful, phased compression in LMM to preserve the model's performance integrity.

\subsection{\textbf{RQ5 \& RQ6}: Distillation Ablation}

\begin{figure}[t]
    \centering
    \includegraphics[width=\linewidth]{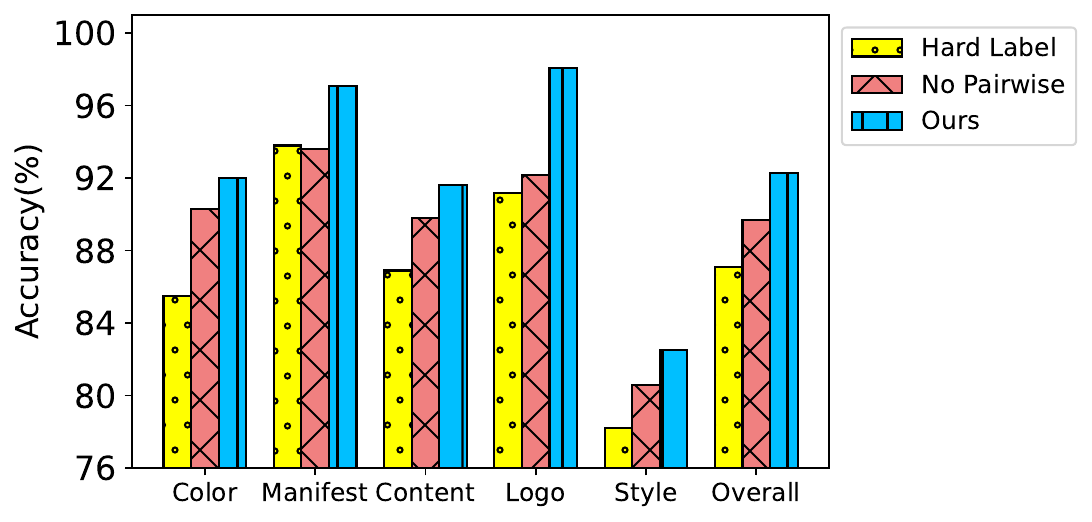} 
    \caption{Efficacy of different distillation techniques on model accuracy. This figure depicts the comparative results of employing various distillation methods.}
    \label{fig:distillation_methods}
    \vspace{-0.4cm}
\end{figure}

\noindent We also compared different distillation methods, as shown in Table~\ref{tab:distillation_methods} and Figure~\ref{fig:distillation_methods}.
We observed that when the pairwise loss was not used, the accuracy achieved was 0.871. However, our proposed method outperformed both approaches, achieving an accuracy of 0.923.
These findings indicate that incorporating the pairwise loss method significantly enhances the distillation process, leading to higher accuracy compared to not using the pairwise loss. Hence, our method demonstrates its effectiveness in improving the performance of the distillation process.
Interestingly, we also explored the utilization of hard labels during the distillation process, where the labels from the dataset are employed as supplementary supervision information. However, our findings revealed that employing solely the outputs of the larger model as soft labels resulted in significant improvements in model performance.

%% file: Content/Related_works.tex
\vspace{-0pt}
\section{Online Improvement}
Compressed AntGMM with our method was deployed online at the end of September. Compared to the direct application of AntGMM, our method witnessed a marginal decrease in the seven-day average accuracy by $0.8\%$; however, it significantly reduced the service latency from approximately \textbf{700ms to 90ms}. In contrast to approaches not employing LMM, our model demonstrated an \textbf{$18\%$} increase in the seven-day average accuracy while incurring a mere 5ms increase in latency.
\vspace{-0pt}
\section{Related Works}

\noindent\textbf{Large Multimodal Model.}
Advancements in LMMs, such as BLIP-2's use of frozen FlanT5 models for visual features extraction~\cite{li2023blip}, MiniGPT-4's integration of a visual encoder with Vicuna LLM and ChatGPT captions~\cite{zhu2023minigpt}, and LLaVA's combination of a visual encoder output with LLaMA~\cite{li2023llava}, have significantly enhanced vision-language tasks. Additionally, mPLUG-Owl introduces a modular training paradigm that surpasses MiniGPT-4 and LLaVA in understanding instructions and visuals~\cite{ye2023mplug}. Our company's current version of AntGMM adopts a structure based on BLIP-2, and we are exploring other approaches and employing various techniques to improve our service quality.

\begin{table}[t]
\centering
\caption{Comparison of Distillation Methods}
\label{tab:distillation_methods}
\begin{tabular}{@{} l S[table-format=1.3] S[table-format=1.3] S[table-format=1.3] @{}}
\toprule
 & {Hard label Distillation} & {No Pair-wise loss} & {Ours} \\
\midrule
Accuracy & 0.871 & 0.897 & 0.923 \\
\bottomrule
\end{tabular}
\end{table}

\noindent\textbf{Large Model Compression.} 
Large model compression techniques encompass network pruning~\cite{frantar2023sparsegpt,sun2023simple,ma2023llmpruner}, knowledge distillation~\cite{liang2023homodistil,shridhar2023distilling,gu2023knowledge}, quantization~\cite{dettmers2022llm,liu2023llm,xiao2023smoothquant}, and are enhanced by strategies like early exit~\cite{del2023skipdecode} and dynamic token reduction~\cite{del2023skipdecode}. Our work zeroes in on language model pruning, with a spotlight on structural pruning and distillation. Structural pruning, which targets the elimination of filters for hardware efficiency, draws on methods such as l1-dependent pruning~\cite{zafrir2021prune}, first-order importance estimation~\cite{hou2020dynabert}, and Hessian-based estimation~\cite{ma2023llmpruner,wang2019eigendamage}. 
In structural pruning, the debate over the pruning unit ranges from entire layers~\cite{fan2019reducing} to more granular structures like multi-head attention~\cite{voita2019analyzing} or feed-forward layers~\cite{mccarley2019structured}. We implement a multi-stage approach, aiming to diminish redundancy at multiple levels.

\vspace{-0pt}

%% file: Content/Appendix.tex
\section{Case Study}

\begin{figure}[H]
    \centering
    \includegraphics[width=\linewidth]{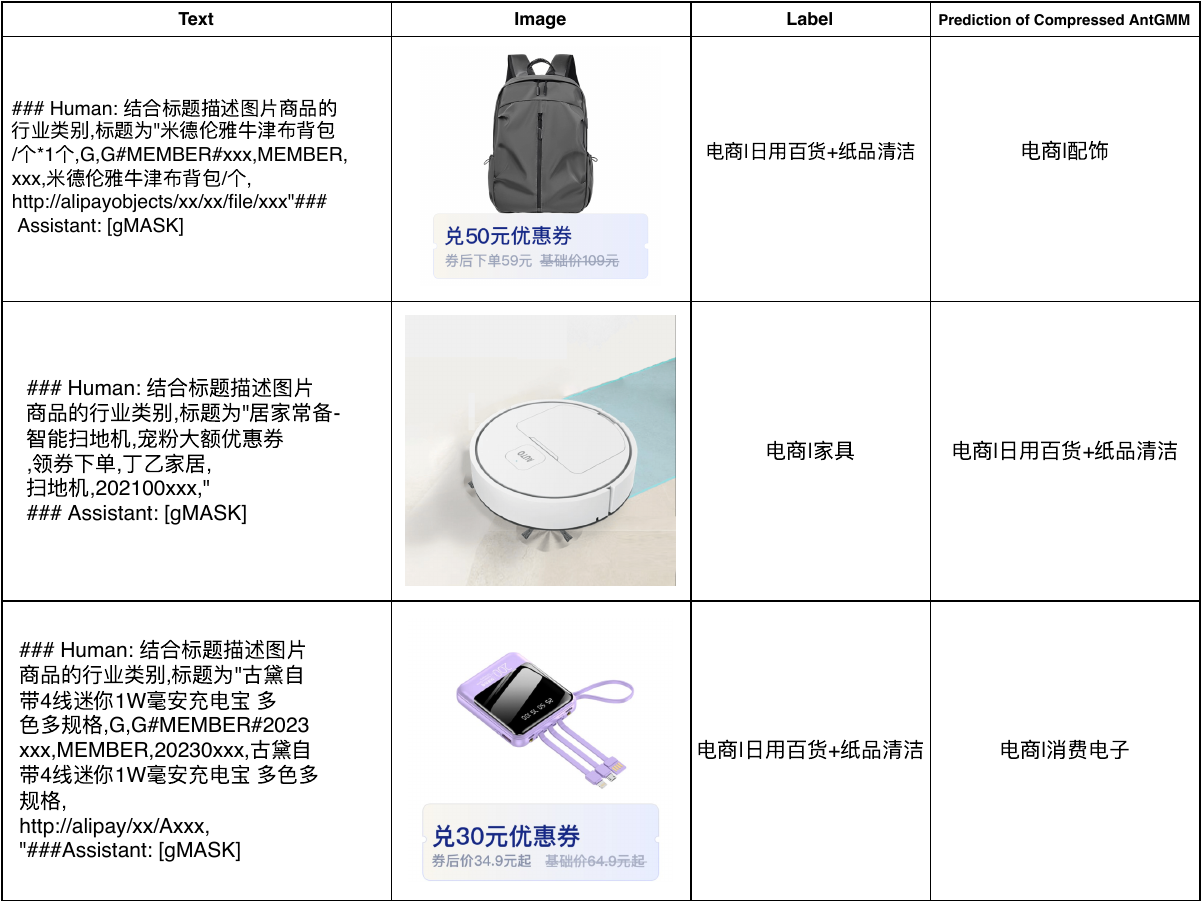} 
    \caption{Some bad caeses. Some sensitive information has been processed.}
    \label{fig:sample}
\end{figure}

As shown in Figure~\ref{fig:sample}, it is noteworthy that we selected several misclassified samples and found that many of the categorization errors are related to our ambiguous labeling, which are also difficult for humans to discern. Nonetheless, it is important to highlight that such classification discrepancies are unlikely to substantially affect real-world business applications.

\section{Model Size and Steps}

\begin{figure}[H]
    \centering
    \includegraphics[width=0.6\linewidth]{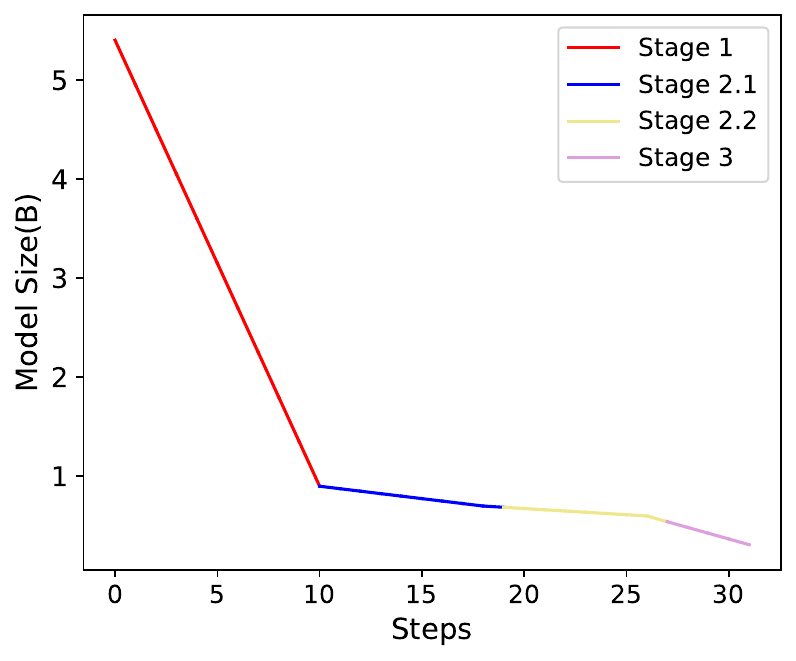} 
    \caption{Correlation between model size and step.}
    \label{fig:modelsize}
\end{figure}

As depicted in the Figure~\ref{fig:modelsize}, we observe the variation in model size with respect to the number of steps. It is evident that the parameter count decreases most rapidly during the block pruning phase. This is followed by a more gradual reduction in the inter-module stage, and a modest decrease is also noted in the input/output parameters.

\section{Model Size and Performance}

\begin{figure}[H]
    \centering
\includegraphics[width=0.6\linewidth]{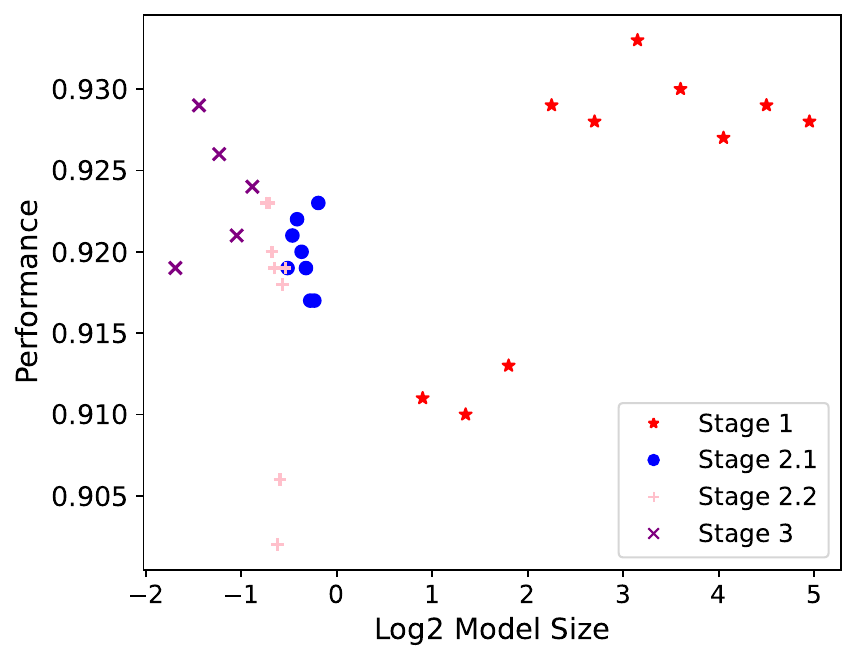} 
    \caption{Correlation between model size and performance.}
    \label{fig:modelsize2}
\end{figure}

The Figure~\ref{fig:modelsize2} illustrates the relationship between model size and performance. To enhance clarity, we have applied a log2 transformation to the values representing model size. Our observations indicate that generally, a smaller model correlates with reduced performance within a given phase. However, there are instances of significant performance improvement, suggesting an absence of a consistent pattern.

\section{Real word display}
\begin{figure}[H]
    \centering
    \includegraphics[width=0.7\linewidth]{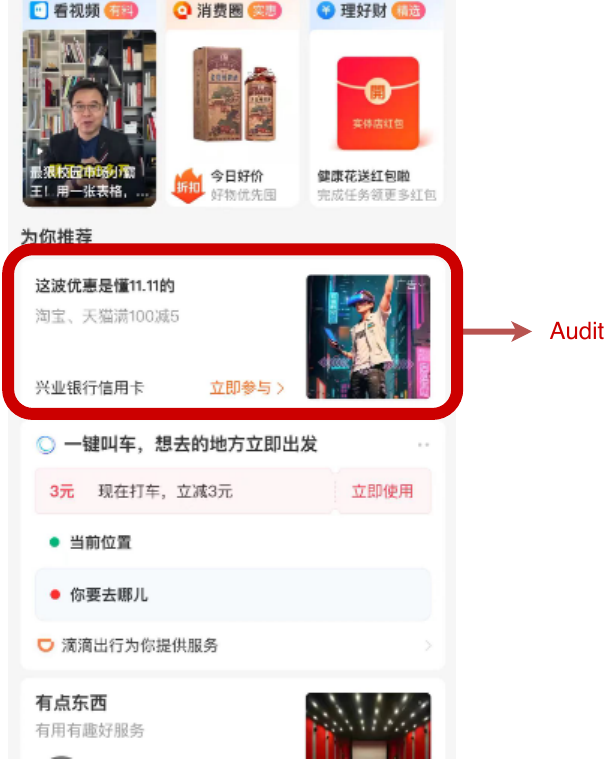} 
    \caption{Real-world Alipay interface. The advertisements to be audited are enclosed within red boxes.}
    \label{fig:realworld}
\end{figure}

Figure~\ref{fig:realworld} showcases real-world multimodal advertisements in Alipay that are subject to auditing.

